\documentclass[10pt,twocolumn,letterpaper]{article}

\usepackage{cvpr}
\usepackage{times}
\usepackage{epsfig}
\usepackage{graphicx}
\usepackage{amsmath}
\usepackage{amssymb}
\usepackage[utf8]{inputenc}
\usepackage{algorithm}
\usepackage{amsfonts}
\usepackage{algpseudocode}
\usepackage{booktabs}
\usepackage{subcaption}

\DeclareMathOperator*{\argmin}{arg\,min}
\DeclareMathOperator*{\softmax}{softmax}
\usepackage{xcolor}

\def\@fnsymbol#1{\ensuremath{\ifcase#1\or \dagger\or \ddagger\or
   \mathsection\or \mathparagraph\or \|\or **\or \dagger\dagger
   \or \ddagger\ddagger \else\@ctrerr\fi}}
   
\newlength\myindent
\usepackage[breaklinks=true,bookmarks=false]{hyperref}

\cvprfinalcopy 

\def\cvprPaperID{6622} 

\ifcvprfinal\fi
\begin{document}

\title{Deep Iterative Surface Normal Estimation}

\author{Jan Eric Lenssen$^{1,2,}$\thanks{Work performed during an internship at NNAISENSE.} \\
\tt\small janeric.lenssen@udo.edu
\and
Christian Osendorfer$^{1}$\\
\tt\small christian@nnaisense.com
\\
\\
$^{1}$ NNAISENSE \\
$^{2}$ TU Dortmund University\\
\and
Jonathan Masci$^{1}$\\
\tt\small jonathan@nnaisense.com
}

\maketitle
\thispagestyle{empty}
\vspace{-1cm}

\begin{abstract}
This paper presents an end-to-end differentiable algorithm for robust and detail-preserving surface normal estimation on unstructured point-clouds. We utilize graph neural networks to iteratively parameterize an adaptive anisotropic kernel that produces point weights for weighted least-squares plane fitting in local neighborhoods.
The approach retains the interpretability and efficiency of traditional sequential
plane fitting while benefiting from adaptation to data set statistics through deep learning.
This results in a state-of-the-art surface normal estimator that is robust to
noise, outliers and point density variation, preserves sharp features
through anisotropic kernels and equivariance through a local quaternion-based spatial transformer. Contrary to
previous deep learning methods, the proposed approach does not require any 
hand-crafted features or preprocessing. It improves on the state-of-the-art results while being more than two orders of magnitude faster and more parameter efficient. 
\end{abstract}
\vspace{-0.3cm}
\section{Introduction}


Normal vectors are local surface descriptors that are used as an input for several 
computer vision tasks ranging from surface reconstruction~\cite{Kazhdan:2006} to 
registration~\cite{Pomerleau:2015} and segmentation~\cite{Grilli:2017}.
For this reason, the task of surface normal estimation has been an important
and well studied research topic for a long time, with 
several methods dating back up to 30 years~\cite{Hoppe:1992}.
Progress in the field, however, has been plateauing only until recently 
when a number of works has shown that improvements can be achieved
with the use of data-driven deep learning \mbox{techniques \cite{Ben-Shabat:2018, Boulch:2016, Guerrero:2018}},
as also shown in related fields like point cloud denoising \cite{Rakotosaona:2019} or finding 
correspondences on meshes and point clouds \cite{Deng:2018b, Fey:2018, Masci:2015, Monti:2017}.
Deep learning methods are known to often achieve better results compared to data-independent methods. However, they have downsides in terms of robustness to small input changes, adversarial attacks, interpretability, and sometimes also computational efficiency. Also, they do not make use of often well-known instrinsic problem structure, which leads to the necessity of having a large amount of training data and model parameters to learn that structure on their own.

 \begin{figure}[t]
\centering
  \includegraphics[width=0.95\linewidth]{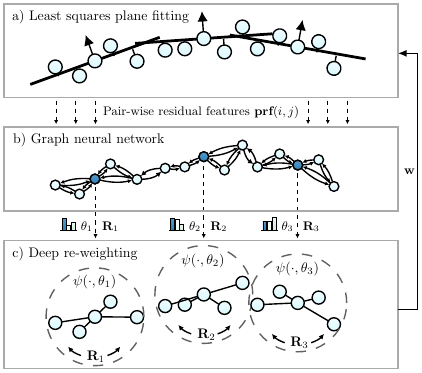}
  \caption{Simplified overview of the proposed method for deep iterative surface normal estimation. The figure shows the process for a subset of three points. (a) Surfaces are fitted by optimizing weighted least squares. (b) A graph neural network infers kernel parameters and local orientation from intermediate pair-wise point descriptors. (c) A trainable, adaptive kernel refines the weights for the next step of the least squares optimization.}
  \label{fig:overview}
\vspace{-0.3cm}
\end{figure}

It is well-known that surface normal estimation can be formulated as a least-squares optimization problem.
A way to utilize this problem-specific knowledge with deep learning is to take an iteratively reweighting least squares (IRLS) scheme \cite{Holland:1977} for robust model fitting and modify it
using deep data-dependent weighting, as it has been done recently (with or without iterations) for other tasks \cite{Huang:2019, Ranftl:2018,Suwajanakorn:2018, Wang:2019}. It is a promising candidate to combine robustness, interpretability and efficiency with the data prior of deep neural networks (DNNs). From a deep learning perspective, the approach imposes a strong bias on the architecture, heavily constraining the space of solutions to those which are better suited for the given problem.

\paragraph{Contribution}
In this work, we present such a trainable re-weighting procedure for input graphs with a large number of weighted least square problems and use it to design a fast and accurate algorithm for surface normal estimation on unstructured point clouds (c.f. Figure \ref{fig:overview}). The method consists of a light-weight graph neural network (GNN), which parameterizes a local quaternion transformer and a deep kernel function to iteratively re-weight graph edges in a large-scale point neighborhood graph. We show that the resulting algorithm 

\begin{itemize}
  \setlength\itemsep{0em}
    \item reaches state-of-the-art performance in surface normal estimation on unstructured point clouds,
    \item is more than two orders of magnitude faster and more parameter efficient than related deep learning approaches, and
    \item is robust to noise and point density variation, while being equivariant and able to preserve sharp features.
    
\end{itemize}

\section{Related work}

Traditional methods for surface normal estimation make use of plane
fitting approaches like unweighted principal component analysis 
(PCA)~\cite{Hoppe:1992} and singular value decomposition (SVD) 
(c.f.~\cite{Klasing:2009} for an overview). 
The performance of these approaches usually hinges upon the often
cumbersome selection of data-specific hyper-parameters, such as
point neighborhood sizes, and it is sensitive to noise, outliers and
density variations.
Because of this, several heuristics have been proposed to ease such selection, e.g. those for finding a neighborhood size
for plane fitting \cite{Mitra:2004}. 
Another limitation of plane fitting methods is that they tend to smoothen sharp details, in fact they can be seen as
isotropic low-pass filters. 
In order to preserve sharp features methods that extract normal vectors 
from estimated Voronoi cells have been 
proposed~\cite{Amenta:1998,Merigot:2011} and combined 
with PCA~\cite{Alliez:2007}. 
Alternative approaches include edge-aware sampling~\cite{Huang:2013} or
normal vector estimation in Hough space~\cite{Boulch:2012}. In
addition, several methods arise from more complex surface reconstruction
techniques, e.g. moving least squares (MLS)~\cite{Levin:1998}, spherical fitting~\cite{Guennebaud:2007}, jet fitting~\cite{Cazals:2003} and multi-scale kernel methods \cite{Aroudj:2017}.

\paragraph{Deep learning methods.} Deep learning based approaches also found their
way into surface normal estimation with the recent success of deep learning in a wide
range of domains. These approaches can be divided into two groups, depending on
the actual type of input data they use. The first group aims at normal estimation 
from single images \cite{Bansal:2016, Eigen:2015, Fouhey:2013, Ladicky:2014, Li:2015, Qi:2018, Wang:2015} and has received a lot of interest over the last few years due to the well understood properties of CNNs for grid-structured data.

The second line of research
directly uses unstructured point clouds and emerged only 
very recently, partially due to the advent of graph neural networks
and geometric deep learning~\cite{BronsteinGDLSurvey:2017}.
Boulch et al.~\cite{Boulch:2016} proposed to use a CNN on
Hough transformed point clouds in order to find surface planes of the
point cloud in Hough space. 
Based on the recently introduced point processing network, 
PointNet~\cite{Qi:2017}, Guerrero et al.~\cite{Guerrero:2018} proposed a 
deep multi-scale architecture for surface normal estimation. 
Later, Ben-Shabat et al.~\cite{Ben-Shabat:2018} improved on those results
using 3D point cloud fisher vectors as input features and a three-dimensional CNN
architecture consisting of multiple expert networks.

\section{Problem and background}
Let $\mathcal{S}$ be a manifold in $\mathbb{R}^3$, $\mathcal{P} = \{\mathbf{p}_0, ..., \mathbf{p}_m$\} a finite set of sampled and possibly distorted points from that manifold and $\hat{\mathbf{N}} = \{\hat{\mathbf{n}}_0, ..., \hat{\mathbf{n}}_m$\} the tangent plane normal vectors at sample points $\mathbf{p}_i$. \emph{Surface normal estimation} for the point cloud $\mathcal{P}$ can be described as the problem of estimating a set of normal vectors $\mathbf{N} = \{\mathbf{n}_0, ..., \mathbf{n}_m\}$ given $\mathcal{P}$, whose direction match those of the actual surface normals $\hat{\mathbf{n}}_i$ as close as possible. We consider the problem of unoriented normal estimation, determining the normal vectors up to a sign flip. Estimating the correct sign can be done in a post-processing step, depending on the task at hand, and is explicitly tackled by several works~\cite{Mullen:2010, Huang:2019, Wu:2015}.

A standard approach to determine unoriented surface normals is fitting planes to the local neighborhood of every point $\mathbf{p}_i$ \cite{Levin:1998}. Given a radius $r$ or a neighborhood size $k$, we model the input as a nearest neighbor graph $G=(\mathcal{P}, \mathcal{E})$, where we have a directed edge $(i,j) \in \mathcal{E}$ if and only if $||\mathbf{p}_j - \mathbf{p}_j||_2 < r$ or if $\mathbf{p}_j$ is one of the $k$ nearest neighbors of $\mathbf{p}_i$, respectively. Let $\mathcal{N}(i)$ denote the local neighborhood of $\mathbf{p}_i$, with $k_i \equiv |\mathcal{N}(i)|$, containing all $\mathbf{p}_j$ with $(i,j) \in \mathcal{E}$. Furthermore, let $\mathcal{P}(i) \in \mathbb{R}^{k_i \times 3}$ be the matrix of centered coordinates of the points from this neighborhood, that is 
\begin{equation}
    \mathcal{P}(i)_j = \mathbf{p}_j^\top - \frac{1}{k_i} \sum_{m\in\mathcal{N}(i)} \mathbf{p}_m^\top \textrm{,}\qquad \mathbf{p}_j \in \mathcal{N}(i).
\end{equation}
Fitting a plane to this neighborhood is then described as finding the least squares solution of a homogeneous system of linear equations:
\begin{equation}
    \mathbf{n}_i^\ast = \argmin_{\mathbf{n}:|\mathbf{n}|=1} 
    || \mathcal{P}(i) \mathbf{n} ||^2_2 =
    \argmin_{\mathbf{n}:|\mathbf{n}|=1} \sum_{j\in\mathcal{N}(i)}  || \mathcal{P}(i)_j\cdot \mathbf{n} ||^2
    \label{eq:plane}
\end{equation}
The simple plane fitting of Eq.~\ref{eq:plane} is not robust and does not result 
in high-quality normal vectors: It produces accurate results only if there are no outliers 
in the data, which is never the case in practice. Additionally, this approach eliminates sharp details because it 
acts as a low-pass filter on the point cloud. 
Even when an isotropic radial kernel function $\theta(||\mathcal{P}(i)||)$ is used
to weight points according to their distance to the local mean, fine details cannot be
preserved.

Both problems can be resolved through integrating weighting functions into Eq.~\ref{eq:plane}. \emph{Sharp features can be preserved} with an \emph{anisotropic} kernel that infers weights of
point pairs based on their relative positions, i.e.: 
\begin{equation}
\label{eq:aniso}
    \mathbf{n}_i^\ast = \argmin_{\mathbf{n}:|\mathbf{n}|=1} \sum_{j\in\mathcal{N}(i)} \psi(\mathbf{p}_j - \mathbf{p}_i) \cdot || \mathcal{P}(i)_j\cdot \mathbf{n} ||^2
\end{equation}
where $\psi(\cdot)$ is an anisotropic kernel, considering the full Cartesian relationship between neighboring points, instead of only their distance. However, an anisotrop kernel is no longer rotation invariant, so that \emph{equivariance} of output normals needs to be ensured additionally. 
\emph{Robustness} to outliers can be achieved by another kernel that weights points 
according to an inlier score $s_{i, j}$. 
More specifically, Eq.~\ref{eq:plane} is changed to 
\begin{equation}
\label{eq:outlier}
    \mathbf{n}_i^\ast = \argmin_{\mathbf{n}:|\mathbf{n}|=1} \sum_{j\in\mathcal{N}(i)} s_{i,j} \cdot || \mathcal{P}(i)_j\cdot \mathbf{n} ||^2 \textrm{,}
\end{equation}
where $s_{i,j}$ weights outliers with a low and inliers with a high score. However, in order to infer information about the outlier status of points an initial model estimation is necessary. A standard solution to this circular dependency is to formulate the problem as a sequence of weighted least-squares problems \cite{Holland:1977, Ranftl:2018}. Given the residuals $\mathbf{r}^l$ of the least squares solution from iteration $l$, the solution for iteration $l+1$ is computed as
\begin{equation}
\label{eq:it_optim}
    \mathbf{n}_i^{l+1} = \argmin_{\mathbf{n}:|\mathbf{n}|=1} \sum_{j\in\mathcal{N}(i)} s(\mathbf{r}^l_{i, j}) \cdot || \mathcal{P}(i)_j\cdot \mathbf{n} ||^2 \textrm{.}
\end{equation}
That is, the inlier score and the estimated model are refined in an alternating fashion.

\section{Deep iterative surface normal estimation}
In this section we present our method, which combines the described properties of robustness, anisotropy and equivariance with the deep learning property of adaptation to large data set statistics. In contrast to existing deep learning methods \cite{Ben-Shabat:2018, Guerrero:2018}, we do not directly regress normal vectors from point features but weights for a least-squares optimization step, utilizing the problem specific knowledge outlined above.

The core of the algorithm is a trainable kernel function $\psi: \mathbb{R}^3 \times \mathbb{R}^d \rightarrow \mathbb{R}$, which computes weights as
\begin{equation}
w_{i,j} = \psi(\mathbf{R}_i(\mathbf{p}_j - \mathbf{p}_i), \theta_i), \label{eq:irlsw}
\end{equation}
where $\theta$ are kernel parameters and $\mathbf{R}$ is a rotation matrix. The kernel is shared by all local neighborhoods of the point graph while $\theta$ and $\mathbf{R}$ are individual for each node. Because there is no apriori information about the structure of the input data, a reasonable approach is to model $\psi$ as an MLP and to find kernel parameters through supervised learning from data.  To this end, parameters $\theta$ and poses $\mathbf{R}$ for each neighborhood are jointly regressed by a graph neural network on the point neighborhood graph. Then, the kernel function $\psi$ regresses anisotropic, equivariant weights $w_{i,j}$ for each edge in the graph, which are used to find the normal vectors using traditional weighted least-squares optimization
\begin{equation}
\mathbf{n}_i = \argmin_{\mathbf{n}:|\mathbf{n}|=1} \sum_{j\in\mathcal{N}(i)} \softmax_{j\in\mathcal{N}(i)}(w_{i,j}) || \mathcal{P}(i)_j\cdot \mathbf{n} ||^2 \textrm{,} \label{eq:irlsn}
\end{equation}
in parallel for all $i \in \mathcal{P}$. 
Similar to iterative re-weighting least squares (c.f. Eq.~\ref{eq:it_optim}), we apply the method in an iterative fashion to achieve robustness and provide the residuals of the previous solution as input to the graph neural network. 

The core algorithm is formulated as pseudo code in Algorithm \ref{alg:main}. The initial weighting of the points in a neighborhood is chosen to be uniform, which results in unweighted least-squares plane fitting in the initial iteration.
In the following, we present the graph neural network, the local quaternion rotation and our differentiable least square solver in more detail.

\begin{algorithm}[t]
\caption{Differentiable iterative normal estimation}
    \label{alg:main}

\begin{algorithmic}
\State \textbf{Input:} 
\State $\mathcal{P}$: Point cloud
\State $L$: Number of iterations 
\State $k$ or $r$: Neighborhood size (num. neighbors or radius)
\State \textbf{Output:} 
\State $\mathbf{N} $: Normal vector estimations
\State --------------------------------------------------------------------
\State $(\mathcal{P},\mathcal{E}) \leftarrow$ Neighborhood graph from $\mathcal{P}$ and $k$ / $r$

 \State  $\mathbf{C} \leftarrow \textrm{CovMatrices}(\mathcal{P}, \mathcal{E})$
 \State  $\mathbf{U}, \mathbf{\Sigma} \leftarrow \textrm{ParallelEig}(\mathbf{C}) $
 \State  $\mathbf{N}^0 \leftarrow$ Extract Solutions from $\mathbf{U}$
 \For{$l \in \{1,...,L\}$} 
  \State $(\Theta, \mathbf{Q}) \leftarrow \textrm{GNN}(\mathcal{P}, \mathcal{E}, \mathbf{N}^{l-1})$ 
  \State $\mathbf{R} \leftarrow \textrm{QuatsToMats}(\mathbf{Q})$
  \State $\mathbf{W} \leftarrow \textrm{ApplyKernel } \psi(\mathbf{R}, \mathcal{P}, \Theta, \mathcal{E})$
  \State  $\mathbf{C} \leftarrow \textrm{WeightedCovMatrices}(\mathcal{P}, \mathbf{W}, \mathcal{E})$
  \State  $\mathbf{U}, \mathbf{\Sigma} \leftarrow \textrm{ParallelEig}(\mathbf{C}) $
  \State  $\mathbf{N}^l \leftarrow$ Extract Solutions from $\mathbf{U}$
\EndFor\\
\Return $\mathbf{N}^L$
\end{algorithmic}
\end{algorithm}

\subsection{GNN for parameterization and rotation}\label{sec:deepw}
For regressing parameters $\mathbf{\theta}$ and rotations $\mathbf{R}$ for the whole point cloud, graph neural networks \cite{Fey:2019, Hamilton:2017} are a natural fit because the network must be invariant to the ordering of the points in a neighborhood and it must be able to allow weight sharing over neighborhoods with varying cardinality.

Our graph neural network architecture consists of a neighborhood aggregation procedure, which is applied three consecutive times. Given MLPs $h$ and $\gamma$, the neighborhood aggregation scheme, similar to that of \mbox{PointNet \cite{Qi:2017}} and to general message passing graph neural network frameworks \cite{Hamilton:2017, Morris:2019}, is given by message function
\begin{equation}
\mathbf{f}_e(i,j) = h\big(\mathbf{f}(i)\,|\,\mathbf{d}_{i,j}\,|\,\mathbf{prf}(i, j)\big) \textrm{,}  
\end{equation}
 and node update function
\begin{equation}
\mathbf{f}(i) = \gamma\Big(\frac{1}{|\mathcal{N}(i)|}\sum_{j\in \mathcal{N}(i)} \mathbf{f}_e(i,j)\Big) \textrm{,} 
\end{equation}
 with $|$ denoting feature concatenation. Using this scheme, we alternate between computing new edge features $\mathbf{f}_e(i,j)$ and node features $\mathbf{f}(i)$. 
 In addition to the Cartesian relation vector \mbox{$\mathbf{d}_{i,j} = (\mathbf{p}_j - \mathbf{p}_i)$}, pair-wise residual features, a modified version of \emph{Point Pair Features (PPF)} \cite{Deng:2018b, Deng:2018a}, are provided as input:
\begin{equation}
    \mathbf{prf}(i, j) = (|\mathbf{n}_i \cdot \mathbf{d}_{i,j}|, |\mathbf{n}_j \cdot \mathbf{d}_{i,j}|, |\mathbf{n}_i \cdot \mathbf{n}_j|, ||\mathbf{d}_{i,j}||_2^2) \textrm{.}
\end{equation}
They are computed directly from the last set of least-squares solutions and contain the residuals as point-plane distances $|\mathbf{n}_i \cdot \mathbf{d}_{i,j}|$. 

After applying the message passing scheme, the output node feature matrix $\mathbf{F} \in \mathbb{R}^{N\times(d+4)}$ is interpreted as a tuple $(\Theta \in \mathbb{R}^{N \times d}, \mathbf{Q} \in \mathbb{R}^{N\times 4})$, containing kernel and rotation parameters for all nodes. We use the row-normalized $\mathbf{Q}$ as unit quaternions to efficiently parameterize the rotation group $SO(3)$. We found that using a rotation matrix instead of an arbitrary $3\times3$ matrix (as in the \emph{Spatial Transformer Network} \cite{Jaderberg:2015}) heavily improves training stability, as also observed by Guerrero et al. \cite{Guerrero:2018}. By applying a custom, differentiable map from quaternion space to the space of rotation matrices we efficiently compute the local rotation matrices $\mathbf{R}$ for all nodes in parallel.

All in all, the graph neural network is permutation invariant, can be efficiently applied in parallel on varying neighborhood sizes, and is a local operator. Locality is an advantage which allows the algorithm to be applied on partial point clouds and scans, without relying on global features or semantics.

\subsection{Parallel differentiable least-squares}
In every iteration of the presented algorithm, the plane fitting problem of Eq.~\ref{eq:irlsn} needs to be solved. A standard approach is to utilize the Singular Value Decomposition of the weighted matrix $\textrm{diag}(\sqrt{\mathbf{w}^{l}_i})\mathcal{P}(i)$: Let $\mathbf{U\Sigma V^T}$ be its decomposition, then the column vector of $\mathbf{V}$ corresponding to the smallest singular value is the optimal solution for the given least squares problem \cite{Hartley:2003, Ranftl:2018}. However, $n$ SVDs (for potentially varying matrix sizes) need to be solved in our scenario, one for every neighborhood, which makes this approach prohibitive. A much more efficient approach in this case is to consider the eigendecomposition of the weighted $3 \times 3$ covariance matrix $\mathbf{C}(i) = \mathcal{P}(i)^\top \textrm{diag}(\mathbf{w}^{l}_i) \mathcal{P}(i)$ which has the columns of $\mathbf{V}$ as its eigenvectors \cite{Hartley:2003}. The solution for Eq.~\ref{eq:irlsn} is then the eigenvector associated with the smallest eigenvalue. The computational complexity for the eigendecomposition of this $3 \times 3$ matrix is $O(1)$ and hence for one overall iteration $O(n)$.

Our algorithm is trained end-to-end by minimizing the distance between ground truth normals and the least squares solution, requiring backpropagation through the eigendecomposition. We follow the work of \mbox{Giles \cite{Giles:2008}}:  Given partial derivatives $\partial L/\partial \mathbf{U}$ and $\partial L/\partial \mathbf{\Sigma}$ for eigenvectors and eigenvalues, respectively, we compute the partial derivatives for a real symmetric $3\times 3$ covariance matrix $\mathbf{C}$ as
\begin{equation}
    \frac{\partial L}{\partial \mathbf{C}} = \mathbf{U} \big((\frac{\partial L}{\partial \mathbf{\Sigma}})_{diag} + \mathbf{F} \circ \mathbf{U}^\top  \frac{\partial L}{\partial \mathbf{U}}\big)\mathbf{U}^\top \textrm{,}
\end{equation}
where $\mathbf{F}_{i,j} = (\lambda_j - \lambda_i)^{-1}$ contains inverse eigenvalue differences.
We implemented forward and backward steps for eigendecomposition of a large number of symmetric $3\times 3$ matrices, where we parallelize over graph nodes, leading to an $O(1)$ implementation (using $O(n)$ processors) of parallel least squares solvers. 

\begin{table*}[t]
\small
  \centering
\begin{tabular}{lllllll}
      \toprule
   & Ours ($k=64$, $L=4$) & Nesti-Net \cite{Ben-Shabat:2018} & PCPNet \cite{Guerrero:2018} & HoughCNN \cite{Boulch:2016} & PCA & Jet \cite{Cazals:2003} \\ \midrule
No noise    & \textbf{6.72} & 6.99      & 9.68   & 10.23    &  12.29   &   12.23  \\ 
Noise ($\sigma = 0.00125$) & \textbf{9.95}    & 10.11     & 11.46  & 11.62    &  12.87   &   12.84  \\ 
Noise ($\sigma  = 0.006$)   &   \textbf{17.18}  & 17.63     & 18.26  & 22.66    &  18.38   &   18.33  \\ 
Noise ($\sigma  = 0.012$)   &  \textbf{21.96}   & 22.28     & 22.8   & 33.39    &   27.5  &   27.68  \\ 
Varying Density (Stripes)    &    \textbf{7.73}         & 8.47      & 11.74  & 12.47    &   13.66  &   13.39  \\ 
Varying Density (Gradients)  &    \textbf{7.51}         & 9.00      & 13.42  & 11.02    &   12.81  &   13.13  \\ 
\midrule
Average &    \textbf{11.84}         & 12.41     & 14.56  & 16.9     &  16.25   &   16.29  \\
\bottomrule
\end{tabular}
\caption{Results for unoriented normal estimation. Shown are normal estimation errors in angle RMSE. For PCA and Jet, optimal neighborhood size for average error is chosen. For our approach, we display results for a balanced neighborhood size $k=64$, which improves on the state of the art for all noise levels. Results for different $k$ are shown in Table \ref{tab:normal_result_comp}.} \label{tab:pcpnet_results}
\vspace{-0.3cm}
\end{table*}

\paragraph{Handling numerical instability.} Backpropagation through the eigendecomposition can lead to numerical instabilities due to at least two reasons: \mbox{1) Low-rank} input matrices with two or more zero eigenvalues. 2) Exploding gradients when two eigenvalues are very close to each other and values of $\mathbf{F}$ go to infinity. We apply two tricks to avoid these problems. First, a small amount of noise is added to the diagonal elements of all covariance matrices, making them full-rank. Second, gradients are clipped after the backward step on very large values, to tackle the cases of nearly equal eigenvalues that lead to exploding gradients.

\subsection{Training}
Training is performed by minimizing the Euclidean distance between estimated normals $\mathbf{N}$ and ground truth normals $\hat{\mathbf{N}}$, averaged over all normal vectors in the training set:
\begin{equation}
L(\hat{\mathbf{N}},\mathbf{N}) =  \frac{1}{n} \sum_{i=1}^n \min (||\hat{\mathbf{n}}_i - \mathbf{n}_i||_2, ||\hat{\mathbf{n}}_i + \mathbf{n}_i||_2)  \textrm{,}
\end{equation}
where the minimum of the distances to the flipped or non-flipped ground truth vectors is used. While we also experimented with different angular losses, we found that the Euclidean distance loss still provides the best result and the most stable training.   A loss is computed after each least squares step and the network is trained iteratively by performing a gradient descent step after each iteration of the algorithm. This fights vanishing gradients that occur due to the normalization of vectors in quaternion and eigenvector computations. The weights of our network are shared over iterations, allowing generalization to further iterations.

\section{Experiments}

Experiments were conducted to compare the proposed Differentiable Iterative Surface
Normal Estimation with state-of-the-art methods both quantitatively,
measuring normal estimation accuracy, and qualitatively, on a Poisson
reconstruction and on a transfer learning task.
Section~\ref{sec:dataset} introduces the dataset used to train our model whereas
Section~\ref{sec:expsetup} details the architecture and the protocol
followed in our experiments. Then, qualitative (Section~\ref{sec:quantitative}) and quantitative (Section~\ref{sec:qualitative}) results are presented and an analysis of complexity and execution time (Section~\ref{sec:complexity}) is given.

\begin{table*}[h]
\small
\centering
\begin{tabular}{llllll|lllll}
\toprule
& \multicolumn{5}{c}{Ours $L=4$} & \multicolumn{5}{c}{PCA} \\
\midrule
 Neighborhood size $k$ & 32  &  48  & 64 & 96 & 128 & 32  &  48  & 64 & 96 & 128  \\ 
\midrule
No noise  & \textbf{6.09}  & 6.63  & 6.72 & 6.82 & 7.35 & 9.10  &  9.94 & 10.68 & 11.93 & 12.54  \\ 
Noise ($\sigma = 0.00125$)  & 10.22 & \textbf{9.63} & 9.95 & 10.45 & 9.64 & 11.22 & 11.56  & 12.08 & 12.71  & 12.97\\ 
Noise ($\sigma  = 0.006$)  & 18.17 & 17.36 & 17.18 & 17.03 & \textbf{16.90} & 28.41 & 23.00 & 20.68 & 18.81 & 18.12 \\ 
Noise ($\sigma  = 0.012$)  & 25.17 & 22.40 & 21.96 & \textbf{21.80} & 22.13 & 45.35 & 38.48 & 33.67 & 28.81 & 26.67 \\ 
Varying Density (Stripes)   & \textbf{7.22}  & 7.63  & 7.73  & 7.87  & 8.67 & 10.48  & 11.40  & 12.07  & 13.18  & 14.07\\
Varying Density (Gradients)   & \textbf{6.84} & 7.19 & 7.51 & 7.69 & 8.49 & 9.96 & 10.74 & 11.35 & 12.36 & 13.21\\ 
\midrule
Average    & 12.28 &  \textbf{11.81} & 11.84 & 11.94 & 12.20 & 19.09 & 17.52 & 16.75 & 16.30 & 16.26 \\ 
\bottomrule
\end{tabular}
\caption{Comparison of unoriented normal estimation RMSE between the proposed method and PCA for different neighborhood sizes $k$. It can be seen that our method consistently provides lower errors while being significantly more robust to changes of that parameter, compared to PCA.}
\label{tab:normal_result_comp}
\vspace{-0.3cm}
\end{table*}

\subsection{PCPNet dataset}
\label{sec:dataset}
Our method is trained and validated quantitatively on the PCPNet dataset as
provided by Guerrero et al. \cite{Guerrero:2018}. 
It consists of a mixture of high-resolution scans, point clouds sampled
from handmade mesh surfaces and differentiable surfaces. 
Each point cloud consists of 100k points. 
We reproduce the experimental setup of~\cite{Ben-Shabat:2018, Guerrero:2018}, 
training on the provided split containing 32 point clouds under different
levels of noise. 
The test set consists of six categories, containing four sets with 
different levels of noise (no noise, $\sigma= 0.00125$, $\sigma= 0.0065$ 
and $\sigma= 0.012$) and two sets with different sampling density 
(striped pattern and gradient pattern). We evaluate  unoriented normal estimation, same as the related approaches.
The Root Mean Squared Error (RMSE) on the provided 5k points subset is 
used as performance metric following the protocol of related work, where the
RMSE is first computed for each test point cloud before the results are
averaged over all point clouds in one category.  
Model selection is performed using the provided validation set.

\subsection{Experimental setup and architecture}
\label{sec:expsetup}
The presented graph neural network was implemented using the 
\emph{Pytorch Geometric} library \cite{Fey:2019}. The neural networks 
$h$, $\gamma$ and $\psi$ each consist of two linear layers, with ReLU
non-linearity. A detailed description of the architecture is
presented in the supplemental materials. During training, output weights from the kernel are randomly set to zero with probability of $0.25$. 

It should be noted that despite inheriting the neighborhood size parameter from traditional PCA, it is possible for a network trained on a
specific $k$ to be applied for other $k$ as well. 
This is because all networks can be shared across an arbitrary number of 
points and the softmax function normalizes weights for neighborhoods of
varying sizes. We observed that generalization
across different $k$ only leads to a very small increase in average
error. 
However, to fairly evaluate our method for different $k$, a network is trained for each $k \in \{32, 48, 64, 96, 128\}$.
Trained consists of 300 epochs using the RMSProp optimization 
method. 
All reported test results are given after $4$ re-weighting iterations of
our algorithm. Iterating longer does not show significant improvements. 
Quantitative results over iterations, results for extrapolation over iterations and generalization between different $k$ are presented in the 
supplemental materials. For further realization details, we refer to our implementation, which is available online\footnote{\footnotesize\url{https://github.com/nnaisense/deep-iterative-surface-normal-estimation}}.

 \begin{figure}[t]
\centering
  \includegraphics[width=1\linewidth]{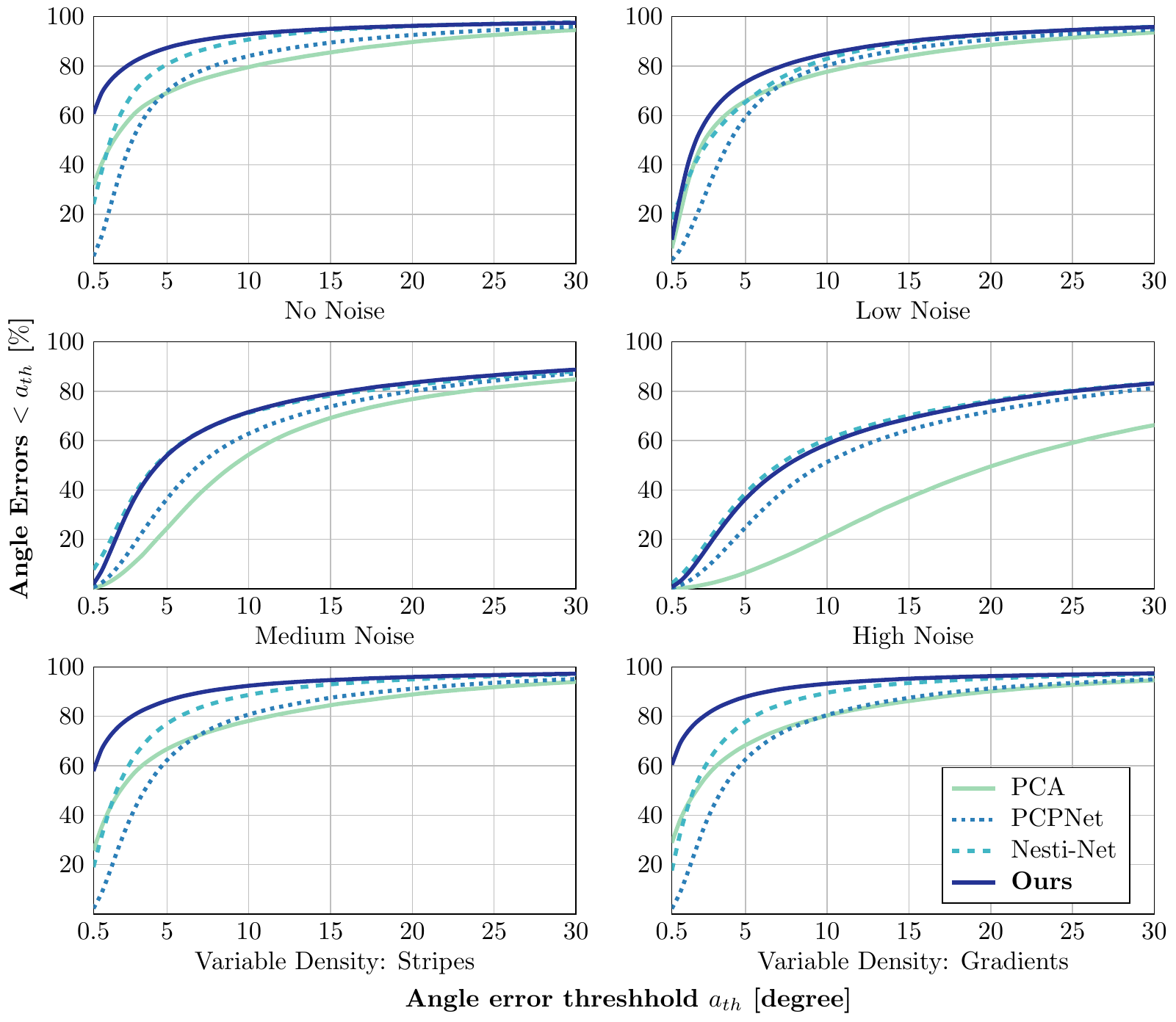}
  \caption{Comparison for varying angle error threshold. For error threshholds on the x-axis, the y-axis shows the percentage of normals which have an error lower than that threshhold. Our method and PCA use neighborhood size $k=64$. For low noise settings and varying density, our method succeeds in recovering sharp features, as shown by the higher accuracies for low angle threshholds.}
  \label{fig:errorplot}
\end{figure}

 \begin{figure*}[t]
\centering
  \includegraphics[width=1\linewidth]{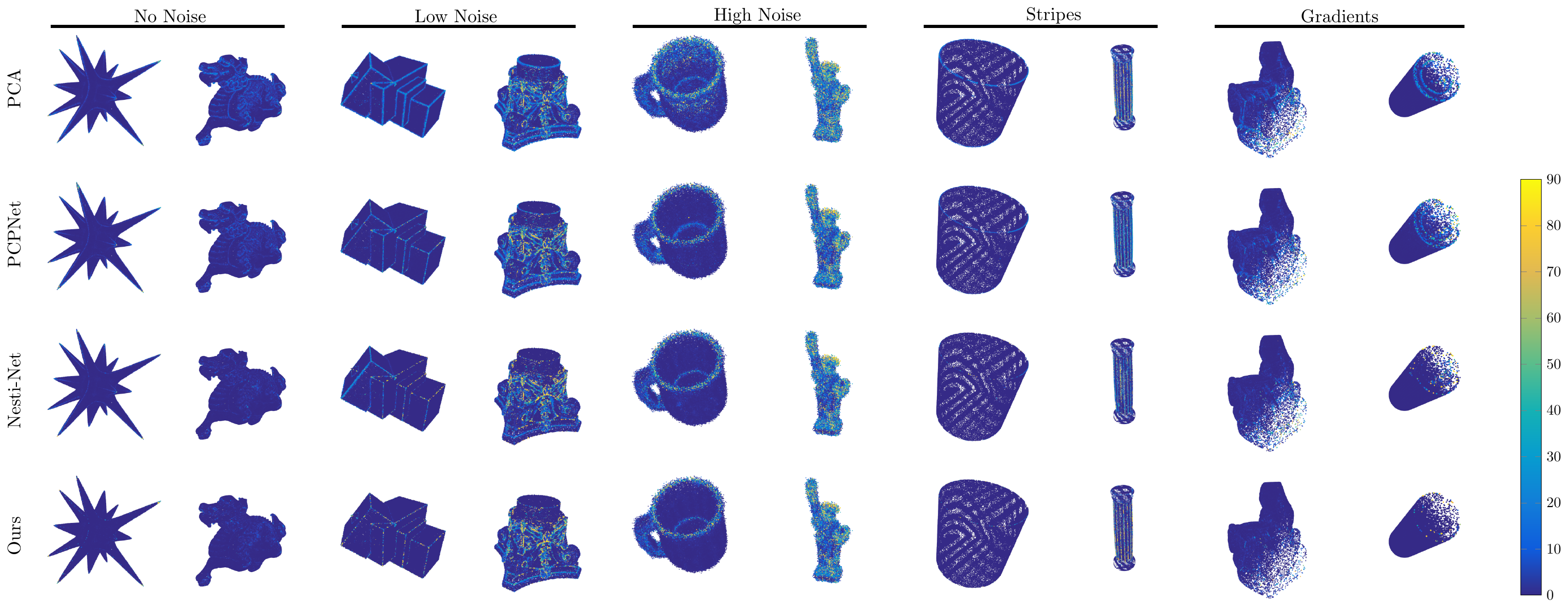}
  \caption{Qualitative comparison between our method ($k=64$, $L=4$) and related work. We show diverse examples from the test set, sampled from different categories, noise levels and density variations. The color encodes the angle error of estimated normals in degrees. Best viewed in the digital version of the paper.}
  \label{fig:qual_comparison}

\end{figure*}

\begin{figure*}[t]
  \begin{subfigure}[t]{0.5\textwidth}
    \includegraphics[width=1\linewidth]{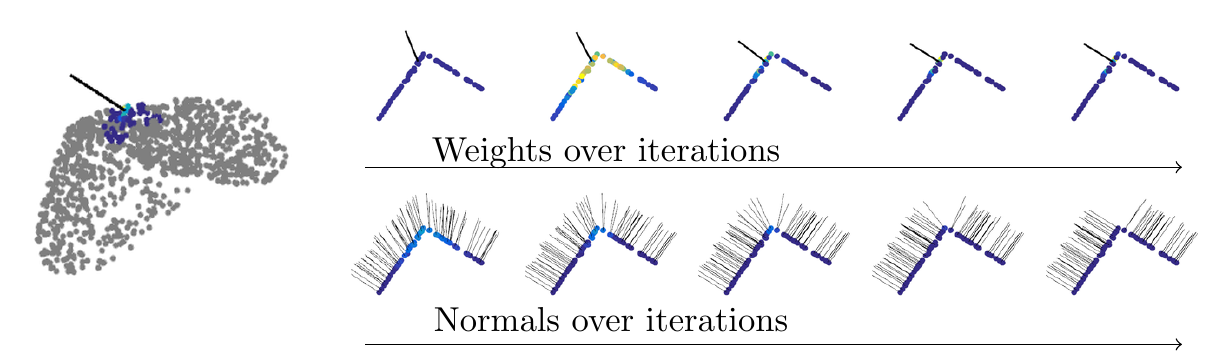}
    \caption{Sharp edge}\label{fig:edge_details}
  \end{subfigure}
  \hfill
  \begin{subfigure}[t]{0.5\textwidth}
    \includegraphics[width=1\linewidth]{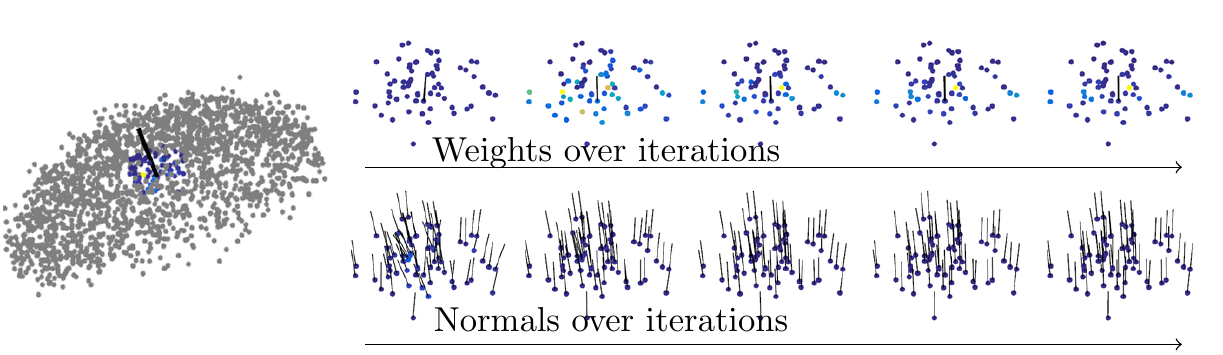}
    \caption{Noisy surface}\label{fig:noisy_details}
  \end{subfigure}
  \caption{Local behaviour of our method over several iterations for a sharp edge (a) and a noisy surface (b). The partial point clouds where sampled from the PCPNet test dataset. The colors in the first rows show the weights from the kernel network for one normal in the neighborhood while the colors in the second row show the angle error of all neighborhood normals.}\label{fig:local_views}

\end{figure*}

\subsection{Quantitative Evaluation}
\label{sec:quantitative}
RMSE results for the PCPNet test set of our approach (with $k = 64$) and related 
works are shown in Table~\ref{tab:pcpnet_results}.  We improve on the
state of the art on all noise levels and varying densities. While the improvement is only small, it should be noted that we reach it while being orders of magnitude faster and more parameter efficient (c.f. Section \ref{sec:complexity}, which is of importance for many applications in resource constraint environments. For the non deep
learning approaches, PCA and Jet, results for medium neighborhood sizes are
displayed. 
In addition, results for different $k$ are provided in 
Table~\ref{tab:normal_result_comp} and compared to errors obtained by PCA with the
same respective neighborhood size. Our method performs stronger than the PCA baseline in all scenarios. 
As expected, varying $k$ leads to a behavior similar to that of PCA, with large 
$k$'s performing better on more noisy data. However, it can be observed that our 
approach is more robust to changes of $k$: Even for small neighborhood sizes, high 
noise is handled significantly better than by PCA and large neighborhoods still 
produce satisfactory results for low noise data. 
It should be noted that for all evaluated $k$ we improve on the state of the
art w.r.t. average error. An evaluation for smaller $k$ down to $k=2$ is provided in the supplemental materials.

While the RMSE error metric is well suited for a general comparison, it is not 
a good proxy to estimate the ability of recovering sharp features since it does not
take into account the error distribution over angles. Therefore, as an additional metric, 
Figure~\ref{fig:errorplot} presents the percentage of angle errors falling below 
different angle thresholds. 
The results confirm that our approach is better at preserving details and sharp
edges, especially for low noise point clouds and varying density, where it 
outperforms other approaches. For higher noise, results similar to 
Nesti-Net are achieved.

\subsection{Efficiency}
\label{sec:complexity}
Our model is small, consisting of only $7981$ trainable parameters,
shared over iterations and spatial locations. 
On a single Nvidia Titan Xp, a point cloud with $100k$ points is 
processed in 5.67 seconds (0.0567 ms per point). 
A large part of this is the kd-tree used to compute the nearest
neighbor graph, which takes 2.1 seconds of the 5.67 seconds.
It is run on the CPU and could be further sped-up by utilizing GPUs.

\begin{table}[h]
\small
  \centering
\begin{tabular}{lrrr}
      \toprule
 & Ours    & Nesti-Net \cite{Ben-Shabat:2018} &  PCPNet \cite{Guerrero:2018}  \\ \midrule
Num. parameters     & \textbf{7981} & $179$M &   $22$M \\ 
Exec. time, 100k p.  & \textbf{3.57 s}  & 1350 s  & 470 s  \\ 
Relative exec. time  & \textbf{1$\times$} &  \textbf{378$\times$}  & \textbf{131$\times$} \\ 
\bottomrule
\end{tabular}
\caption{Comparison of efficiency between the approaches using deep learning. We list number of model parameters as well as average execution times for estimating normals on a point cloud with 100k points.}
\label{tab:complexity}
\end{table}
\vspace{-3mm}

In Table \ref{tab:complexity} we compare our approach against the related deep learning approaches Nesti-Net and PCPNet. Our approach is orders of magnitude (378$\times$ and 131$\times$) faster than the related approaches. The comparison was made as fair as possible by excluding nearest neighbor queries (note that this favors the other approaches since they need larger neighborhoods) and the original implementations.
The speedup of our method can be contributed to the much smaller network size and the parallel design of the GNN and least-squares optimization steps.



\subsection{Qualitative Evaluation}
\label{sec:qualitative}
This section visually presents surface normal errors for various elements of the PCPNet test set in Figure \ref{fig:qual_comparison} and compares them against results from the PCA baseline and related deep learning approaches. It can be seen that the biggest improvements are obtained for low noise scenarios and varying density, where our method is able to preserve sharp features of objects better than the other methods. In general it can be observed that our approach tends to provide sharp, discriminative normals for points on edges instead of smooth averages. In rare cases, this can lead to a false assignment of points to planes, as we can see in the example in column 8. It can be observed  that, in contrast to Nesti-Net, our approach behaves equivariant to input rotation as is seen clearly on the diagonal edge of the box example in column 3. Sharp edges are kept also in uncommon rotations, which we can attribute to our local rotational transformer. Results for more examples are displayed in the supplemental material.

 \begin{figure}[h]
\centering
  \includegraphics[width=1\linewidth]{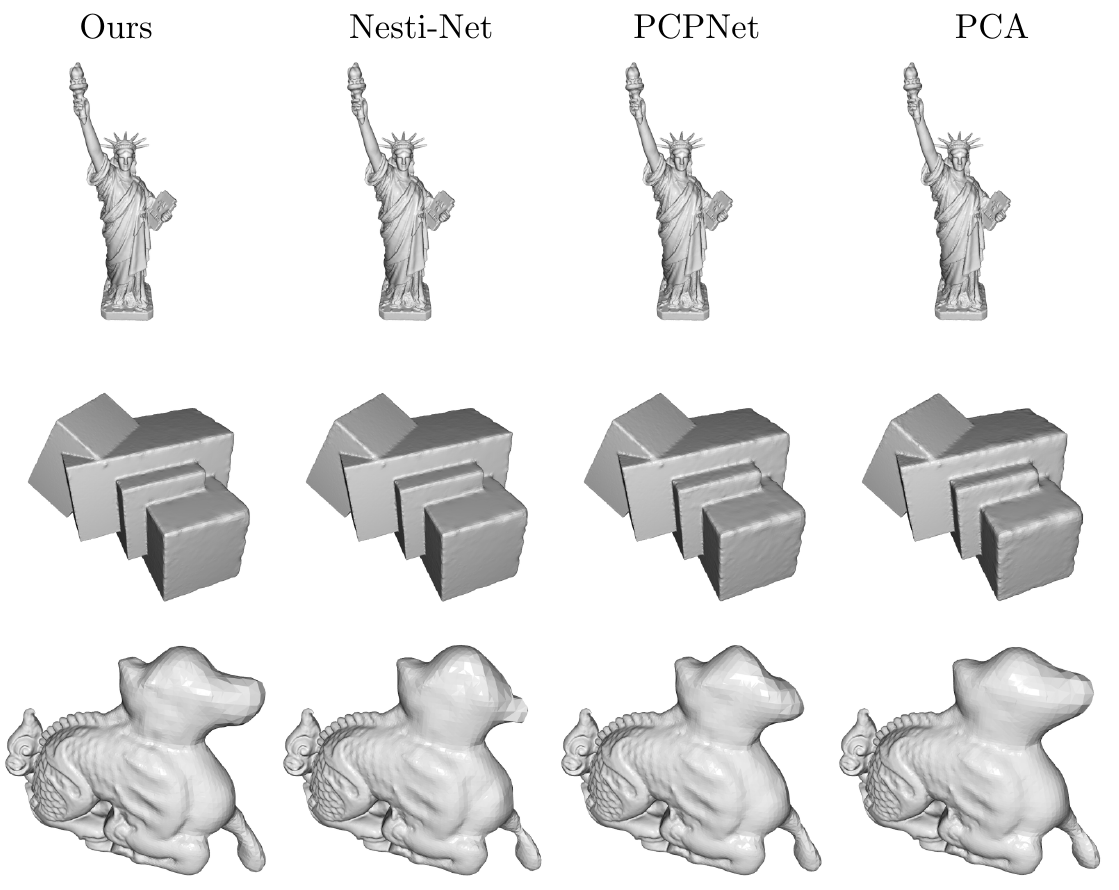}
  \caption{Selected results after applying Poisson surface reconstruction using the estimated normal vectors. In most cases, differences between the methods are very small. Examples 2 and 3 show reconstructions from point clouds with varying density, which show the largest differences.}
  \label{fig:reconstruction}
\end{figure}

\paragraph{Interpretability.}
In order to interprete the results of our method, Figure \ref{fig:local_views} shows a detailed view of local neighborhoods over several iterations of our algorithm. An example for a sharp edge is shown in Figure \ref{fig:edge_details} and a high noise surface in Figure \ref{fig:noisy_details}. Both sets of points were sampled from the real test data. For the sharp edge, the algorithm initially fits a plane with uniform weights, leading to smoothed normals. Over the iterations, high weights concentrate on the more plausible plane, leading to recovering of the sharp edge. In the noisy example, we can see that outliers are iteratively receiving lower weights, leading to more stable estimates.

\paragraph{Surface reconstruction.}
To further evaluate the quality of the produced normals when used
as input to other computer vision pipelines, Figure~\ref{fig:reconstruction}
shows the results for Poisson surface reconstruction. Since the methods in this comparison all perform unoriented normal estimation (Guerrero et al.~\cite{Guerrero:2018} evaluates both, unoriented and oriented, where we chose the unoriented version for a fair comparison), we determine the signs of the output normals from all four methods using the ground truth normals. 
Most of the reconstructions show only small differences, with our
approach and Nesti-Net retaining slightly more details than the others.
Significant differences can be observed for point clouds with varying density,
displayed in rows 2 and 3. Here, our approach successfully retains the 
original structure of the object while still providing sharp edges.

\begin{figure}[t]
\centering
  \includegraphics[width=0.95\linewidth]{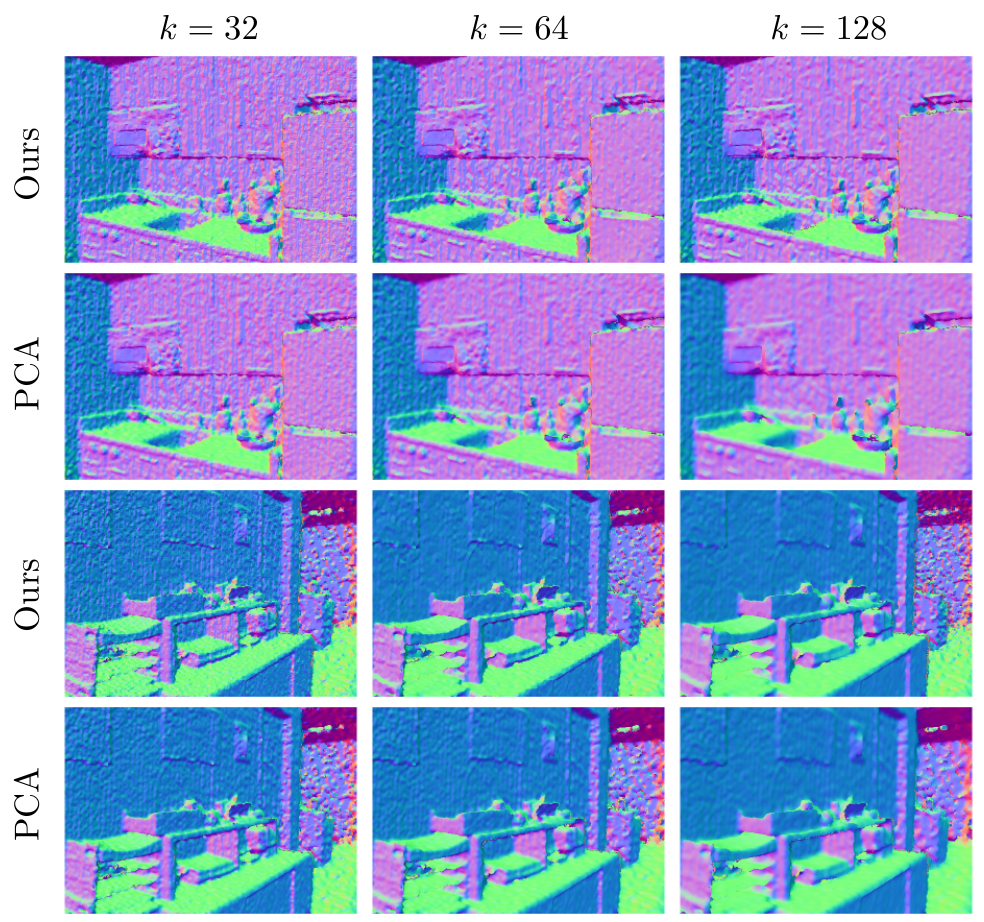}
  \caption{Examples for normal estimation on scanned data from the NYU depth v2 dataset. Colors encode the orientation of normals. Our model generalizes to this dataset while being able to retain more details and sharper edges than PCA. However, scanning artifacts are also kept and visible. Best viewed in the digital version of the paper.
  }
  \label{fig:nyu}
\end{figure}

\paragraph{Transfer to NYU depth dataset.} 
In order to show generality of our approach, our models trained on the PCPNet dataset are validated on the NYU depth v2 dataset \cite{Silberman:2012}, a common benchmark dataset in the field of estimating normals from single images. It contains 1449 aligned and preprocessed RGBD frames, which are transformed to a point cloud before applying our method. After performing unoriented estimation, the normals are flipped towards the camera position.
Evaluation is done qualitatively, since the dataset does not contain ground truth normal vectors. 
Results for three different neighborhood sizes in comparison to PCA are shown in Figure \ref{fig:nyu}. Our approach behaves as expected, as it is able to infer plausible normals for the given scenes. For all $k$, our approach is able to preserve sharp features while PCA produces very smooth results. However, this also leads to the sharp extraction of scanning artifacts, which can be seen on the walls of the scanned room.

\section{Conclusion and future work}
We presented a novel method for deep surface normal estimation on unstructured point clouds, consisting of parallel, differentiable least-squares optimization and deep re-weighting. In each iterations, the weights are computed using a kernel function that is individually parameterized and rotated for each neighborhood by a task-specific graph neural network. The algorithm is much more efficient than previous deep learning methods, reaches state-of-the-art accuracy, and has favorable properties like equivariance and robustness to noise. For future work, investigating the possibility of utilizing deep data priors to parameterize least-squares problems holds large potential. We suspect that introducing data-dependency to other traditional methods can lead to progress in other fields of research, by reducing common disadvantages of pure deep learning approaches. On the theoretical side, it is interesting to dive deeper into convergence properties of IRLS with deep re-weighting. 

\subsection*{Acknowledgements}
We thank Matthias Fey for his work on Pytorch Geometric. We also thank him and Prof. Dr. Heinrich M\"uller for helpful advice and discussions.

{\small
\bibliographystyle{ieee_fullname}
\bibliography{egbib}
}

\appendix
\section*{Supplemental Materials}
\section{Overview}
The supplemental materials contain details about the graph neural network in Section \ref{sec:architecture}, information about the implementation in Section \ref{sec:implementation}, a short discussion about the spatial transformer in \ref{sec:transformer}, and an additional analysis of accuracy over re-weighting iterations in Section \ref{sec:iterations}. Further, we show results for transferring models between different neighborhood sizes in Section \ref{sec:transfer} and qualitative results for the whole PCPNet test set in Section \ref{sec:qualitative}.

\section{Architecture Details}
\label{sec:architecture}
The graph neural network for the deep kernel parameterization follows a general message passing scheme \cite{Fey:2019} with edge update function
\begin{equation}
\mathbf{f}_e(i,j) = h\big(\mathbf{f}(i)\,|\,\mathbf{d}_{i,j}\,|\,\mathbf{prf}(i, j)\big)
\end{equation}
and node update function
\begin{equation}
\mathbf{f}(i) = \gamma\Big(\frac{1}{|\mathcal{N}(i)|}\sum_{j\in \mathcal{N}(i)} \mathbf{f}_e(i,j)\Big) \textrm{,} 
\end{equation}

consisting of 6 MLPs, $h_i$ and $\gamma_i$ for $i \in \{1,2,3\}$. Together with the kernel MLP $\psi$, all functions are detailed in Table~\ref{tab:architecture_details}
The $h_i$ and $\psi$ networks are shared over all edges in the neighborhood graph while the $\gamma_i$ are shared over all points. Additionally, all MLPs are  shared over the iterations of the algorithm. Each MLP consists of two linear layers, seperated by a ReLU non-linearity. Layer sizes are given in \mbox{Table \ref{tab:architecture_details}}. All in all, the networks contain $7981$ parameters and fulfill the following properties.
\begin{table}[h]
  \centering
\begin{tabular}{l|rcl}
      \toprule
                                     Network     &   \multicolumn{3}{c}{Architecture} \\ \midrule
$h_1$   &  $ L(32)$, & $ReLU $, &  $L(16) $ \\ 
$\gamma_1$   &  $L(32)$, & $ReLU $, &  $ L(8) $ \\ 
$h_2$ &  $ L(32)$, &  $ ReLU $, &  $ L(16)$  \\ 
$\gamma_2$   &  $ L(32) $, &  $ ReLU $, &  $ L(8) $ \\ 
$h_3$   & $ L(32) $, &  $ ReLU $, &  $ L(16)$   \\
$\gamma_3$   & $ L(32) $, &  $ ReLU $, &  $ L(12) $  \\ 
$\psi$   & $ L(64) $, &  $ ReLU $, &  $ L(1) $  \\
\bottomrule
\end{tabular}
\caption{Details of the used graph neural network for iterative re-weighting. $L(x)$ stands for a fully-connected layer with $x$ output neurons.} \label{tab:architecture_details}
\end{table}

\paragraph{Permutation Invariance}
Neighborhood aggregation is performed using an average operator, which is invariant regarding the order of points. Since there are no other functions over sets of points, the resulting network is permutation invariant. We refer to \cite{Qi:2017} for further discussion. It should be noted that PointNet can also be expressed in the same message passing scheme and is permutation invariant for the same reasons.

\begin{figure*}[t]
\centering
  \includegraphics[width=0.75\linewidth]{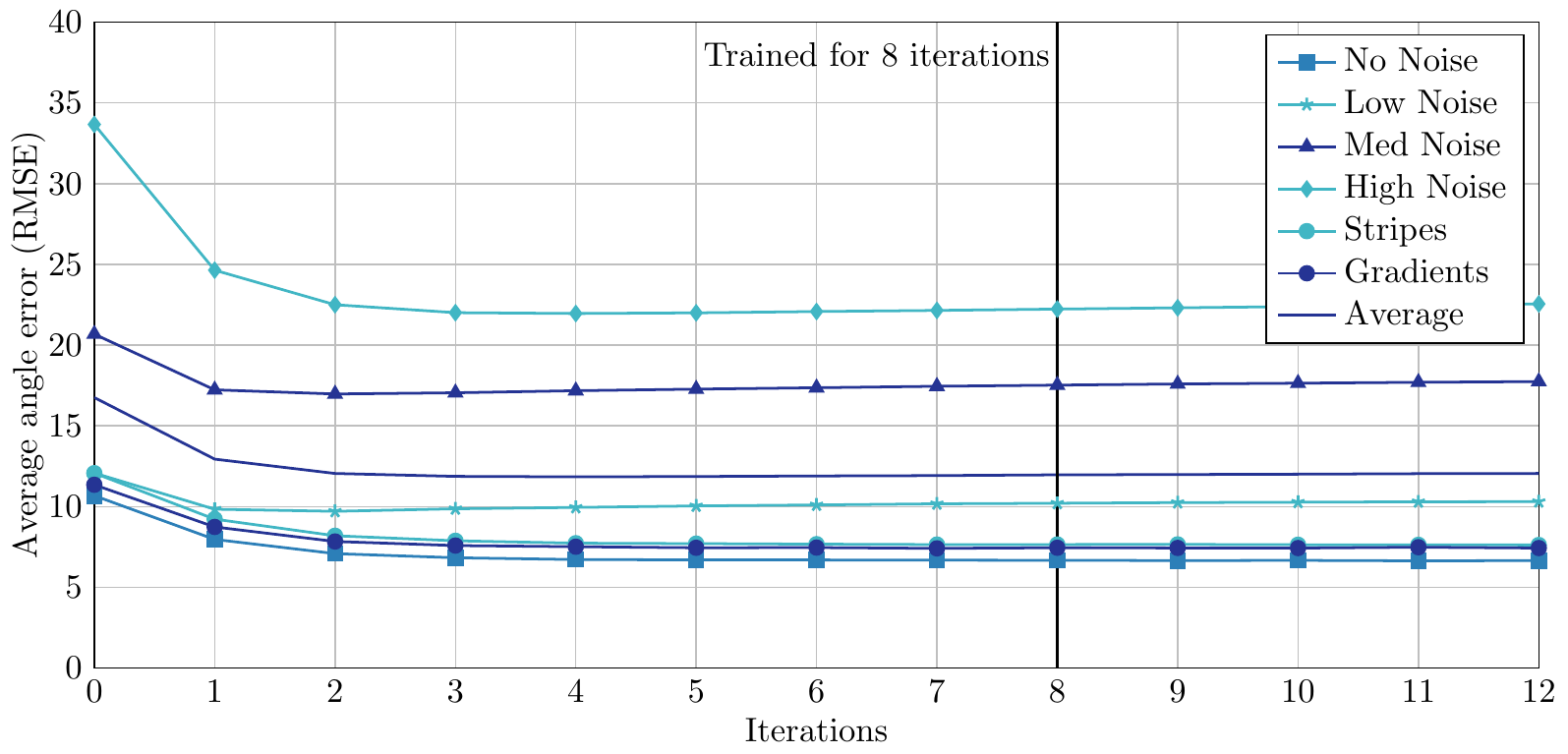}
  \caption{Test errors (RMSE) over iterations of the proposed algorithm. Iteration $0$ shows results for unweighted PCA only. The network was trained on the training set for 8 iterations. For evaluation, we perform four additional iterations to evaluate stability.}
  \label{fig:iterations}
\end{figure*}

\begin{table*}[t]
\small
\setlength\tabcolsep{3.5pt}
\centering
\begin{tabular}{llllll|lllll|lllll}
\toprule
& \multicolumn{5}{c}{Trained on $k^{\textrm{train}} = 32$} & \multicolumn{5}{c}{Trained on $k^{\textrm{train}} = 64$} & \multicolumn{5}{c}{Trained on $k^{\textrm{train}} = 128$} \\
\midrule
$k^{\textrm{test}}$ & 32  &  48  & 64 & 96 & 128 & 32  &  48  & 64 & 96 & 128 & 32  &  48  & 64 & 96 & 128  \\ 
\midrule
No noise  & 6.09 &  6.96 & 7.43 & 8.25 & 8.77 & 6.13  & 6.47 & 6.72 & 7.10 & 7.27 & 6.66 &  7.01 & 7.24 & 7.29 & 7.35 \\ 
Noise ($\sigma = 0.00125$)  & 10.22 & 10.01  & 10.09 & 10.37 & 10.62 & 10.19 &  9.93 & 9.95 & 10.18  & 10.35 & 9.89  & 9.57 & 9.50 & 9.50 & 9.64 \\ 
Noise ($\sigma = 0.006$)   & 18.17 & 17.44 & 17.22 & 17.08 & 17.05 & 18.28 & 17.43 & 17.18 & 17.01 & 16.94 & 20.98 & 18.40 & 17.63 & 17.07 & 16.90 \\ 
Noise ($\sigma = 0.012$) & 25.17 & 22.97 & 22.33 & 21.91 & 21.80 & 25.20 & 22.53 & 21.96 & 21.69  & 21.67 & 30.99 & 24.94 & 23.20 & 22.34 &  22.13 \\ 
Density (Stripes)   &  7.22 & 7.92 & 8.51  & 9.43 & 9.90 & 7.21  & 7.55  &  7.73 & 8.16  &  8.34 &  7.80 & 8.14 &  8.37 & 8.61  & 8.67\\
Density (Gradients)   & 6.84 & 7.46 & 8.06 & 8.80 & 9.21 & 6.89 & 7.17 & 7.51 & 8.04  & 8.03 & 7.48 & 7.75 & 8.11 & 8.39 & 8.49 \\ 
\midrule
Average    & 12.28 & 12.12  & 12.27 & 12.64 & 12.89 & 12.31 & 11.85 & 11.84 & 12.00  & 12.10 & 13.97 & 12.63  & 12.34 & 12.20 & 12.20 \\ 
\bottomrule
\end{tabular}
\caption{Results for transferring models between different neighborhood sizes $k$. Shown are RMSE values for models trained with $k^{\textrm{train}} \in \{32, 64, 128\}$, each tested with $k^{\textrm{test}} \in \{32, 48, 64, 96, 128\}$.}
\label{tab:transfer}
\end{table*}

\begin{table}[t]
\small
\setlength\tabcolsep{3.5pt}
\centering
\begin{tabular}{lllllll}
\toprule
& \multicolumn{6}{c}{Trained on $k^{\textrm{train}} = 32$} \\
\midrule
$k^{\textrm{test}}$ & 2  &  4  & 8 & 16 & 24 & 32  \\ 
\midrule
No noise  & 17.26 &  7.23 & 5.63 & 5.36 & 5.77 & 6.09  \\ 
Noise ($\sigma = 0.00125$)  & 54.02 & 49.66  & 33.65 & 13.80 & 10.74 & 10.22 \\ 
Noise ($\sigma = 0.006$)   & 61.08 & 60.91 & 55.32 & 28.17 & 19.78 & 18.17  \\ 
Noise ($\sigma = 0.012$) & 61.29 & 61.26 & 58.89 & 41.37 & 28.99 &  25.17  \\ 
Density (Stripes)   &  19.50 & 8.14 &6.53 & 6.36 & 6.71 & 7.22 \\
Density (Gradients)   & 22.89 & 8.44 & 6.51 & 6.23 & 6.57 & 6.84 \\ 
\midrule
Average    & 39.34 & 32.59 & 27.75 & 16.88 & 13.09 & 12.28 \\ 
\bottomrule
\end{tabular}
\caption{Results for transferring  the model trained on $k^{\textrm{train}} = 32$ to even smaller $k^{\textrm{test}} \in \{2, 4, 8, 16, 24, 32\}$ until the method breaks down. Note that $k^{\textrm{test}} = 2$ means $2$ neighbors, excluding point $i$, so there are still $3$ points in total for each neighborhood, avoiding underdefined plane fitting problems.}
\label{tab:transfer_smallk}
\end{table}

\paragraph{Varying neighborhood sizes} For the cases in which we decide to use a radius graph instead of a k-nn graph, the network allows differently sized neighborhoods in one graph, since all parameters are shared over edge or nodes and the only operation over the whole neighborhood, the average, is agnostic to the neighborhood size.

\paragraph{Locality}

Due to using only local operators, the presented algorithm can be applied on partial point clouds, which is of importance for many practical applications.

\section{Implementation Details}
\label{sec:implementation}
The implementation of the proposed algorithm is based on the \emph{Pytorch Geometric} library \cite{Fey:2019} and uses the provided scheme consisting of scattering and gathering between node and edge feature space. Therefore, varying neighborhood sizes (\eg varying node degree) can still be handled in parallel on the GPU by parallelization in graph edge space.

For parallel eigendecomposition of a large number of symmetric $3\times3$ matrices and for the parallel quaternion to rotation matrix map, we provide our own Pytorch extensions which is available online. We provide efficient forward and backward steps on GPU and CPU.

\section{Rotational Spatial Transformer}
\label{sec:transformer}
Our spatial transformer learns to bring the point sets in canonical orientation, which leads to equivariant behaviour, as our results show. Directly parameterizing $3\times 3$ matrices for the spatial transformer would lead to arbitrary affine transformations which
can easily collapse or diverge during training. Thus, parameterizing the rotation group $SO(3)$ is the more fitting choice for the given
task. Unit quaternions are a good representation choice because they cover $SO(3)$ (twice) without any discontinuities, as exist in e.g. Euler angles or axis-angle representations. Discontinuities in the $SO(3)$ representation would
force the network to sometimes predict very different values for $SO(3)$ elements that lie next to each other on the
Lie group manifold, which can lead to unstable gradients.

\begin{figure*}[t]
\centering
  \includegraphics[width=1.\linewidth]{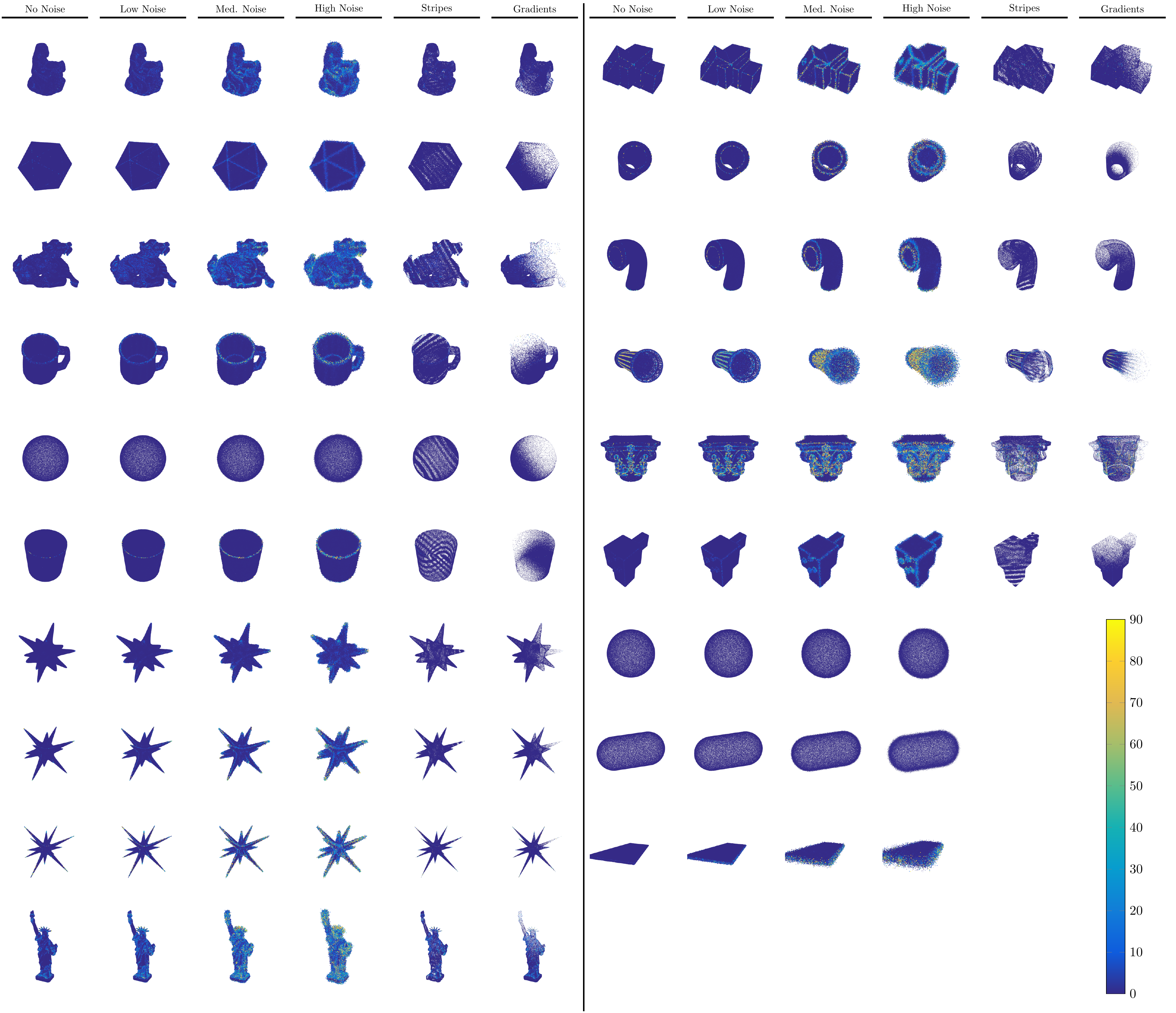}
  \caption{Qualitative results for all examples of the test set. Colors encode the RMSE in degree for each point. Best viewed in the digital version.}
  \label{fig:qualitative}
\end{figure*}

\section{Behaviour over iterations}
\label{sec:iterations}
The algorithm is trained for $L=8$ (performing $8$ iterations of re-weighting), where we compute a loss and perform an optimization step after each iteration. It produces normal vector estimations after each iteration, which can be analyzed quantitatively. The RMSE results for the PCPNet test set over algorithm iterations are shown in Figure \ref{fig:iterations}. It can be seen that after iteration 4, further iterations do not lead to significant improvements. Also, the algorithm behaves reasonable stable, not diverging immediately after we pass the iterations for which the network was trained. However, we observe a small drift in favor of low-noise datasets over the iterations. Errors for the test sets with no noise or variable density still decrease further while errors for data with higher noise levels slightly increase. Meanwhile, the average error stays nearly constant.

\section{Transfer between neighborhood sizes}
\label{sec:transfer}
As stated in the main paper, the proposed algorithm generalizes reasonably well between neighborhood sizes, meaning that a model trained using neighborhood size $k^{\textrm{train}}$ can be applied using a different neighborhood size $k^{\textrm{test}}$ while producing good results. For verification, we report RSME errors for different combinations of $k^{\textrm{train}}$ and $k^{\textrm{test}}$ in Table \ref{tab:transfer}. It can be seen that if the difference in neighborhood size is not too big, transferred models often only perform slightly worse than models trained directly for the appropriate $k$. However, transferring over very large difference like from $128$ to $32$ or the other way around, leads to a significant decrease in performance. The model trained on the balanced $k=64$ performs very well on all other neighborhood sizes.

Additionally, Table \ref{tab:transfer_smallk} provides results for applying the model on even smaller neighborhood sizes, to evaluate the minimum $k$ before the method breaks down. We found that when using a $k^{\textrm{train}} < \approx 30$, the training becomes unstable, which is why we transfer the model from $k^{\textrm{train}} = 32$ to smaller $k^{\textrm{test}} = 32$. Results show that the algorithm provides good results for noise-free data down to $k=4$. For noisy data, the approach breaks down quite fast when lowering $k$, as expected: At least $k=24$ is required to provide reliable results. For lower $k$, the results approach the accuracy of random normals.

\section{Further qualitative results}
\label{sec:qualitative}
\vspace{-0.1cm}
Last, we provide qualitative results for the whole PCPNet test set in Figure \ref{fig:qualitative}. For point clouds with varying density, the point size is reduced in order to better visualize the densities. Similar to examples shown in the paper, we can see that the method produces very sharp normal vectors, which usually resemble the plane normal of one of the plausible planes in the neighborhood. The abstract objects are good examples to show equivariance, as all edges show similar errors, independent of orientation. Sometimes, points are assigned to a false plane, leading to high error normal vectors. Compared to other approaches, we do not observe heavy smoothing around edges.

\end{document}